\documentclass[preprint,12pt]{elsarticle}

\usepackage{amssymb}
\usepackage{amsmath}
\usepackage{graphicx}
\usepackage{subcaption}
\usepackage{multirow}
\usepackage{booktabs}
\usepackage{algorithm}
\usepackage{algorithmic}

\journal{Applied Soft Computing}

\begin{document}

\begin{frontmatter}

%% Title, authors and addresses

%% use the tnoteref command within \title for footnotes;
%% use the tnotetext command for theassociated footnote;
%% use the fnref command within \author or \affiliation for footnotes;
%% use the fntext command for theassociated footnote;
%% use the corref command within \author for corresponding author footnotes;
%% use the cortext command for theassociated footnote;
%% use the ead command for the email address,
%% and the form \ead[url] for the home page:
%% \title{Title\tnoteref{label1}}
%% \tnotetext[label1]{}
%% \author{Name\corref{cor1}\fnref{label2}}
%% \ead{email address}
%% \ead[url]{home page}
%% \fntext[label2]{}
%% \cortext[cor1]{}
%% \affiliation{organization={},
%%             addressline={},
%%             city={},
%%             postcode={},
%%             state={},
%%             country={}}
%% \fntext[label3]{}

% \title{Adaptive Step Size Quantization: An Effective Learnable Adaptive Neural Network Quantization Method}
\title{Precision Neural Network Quantization via Learnable Adaptive Modules}

%% use optional labels to link authors explicitly to addresses:
%% \author[label1,label2]{}
%% \affiliation[label1]{organization={},
%%             addressline={},
%%             city={},
%%             postcode={},
%%             state={},
%%             country={}}
%%
%% \affiliation[label2]{organization={},
%%             addressline={},
%%             city={},
%%             postcode={},
%%             state={},
%%             country={}}

\author[a,b]{Wenqiang Zhou} %% Author name
\author[a,b]{Zhendong Yu} %% Author name
\author[a,b]{Xinyu Liu}
\author[a,b]{Jiaming Yang}
\author[a,b]{Rong Xiao}
\author[a,b]{Tao Wang}
\author[a,b]{Chenwei Tang\corref{*}}
\author[a,b]{Jiancheng Lv}
%% Author affiliation
\affiliation[a]{organization={College of Computer Science, Sichuan University},%Department and Organization
            % addressline={}, 
            city={Chengdu},
            postcode={610065}, 
            % state={},
            country={China}}

\affiliation[b]{organization={Engineering Research Center of Machine Learning and Industry Intelligence, Ministry of Education},%Department and Organization
            % addressline={}, 
            city={Chengdu},
            postcode={ 610065}, 
            % state={},
            country={China}}
\cortext[*]{Corresponding author: tangchenwei@scu.edu.cn}

%% Abstract
\begin{abstract}
%% Text of abstract
Quantization Aware Training (QAT) is a neural network quantization technique that compresses model size and improves operational efficiency while effectively maintaining model performance. The paradigm of QAT is to introduce fake quantization operators during the training process, allowing the model to autonomously compensate for information loss caused by quantization. Making quantization parameters trainable can significantly improve the performance of QAT, but at the cost of compromising the flexibility during inference, especially when dealing with activation values with substantially different distributions. In this paper, we propose an effective learnable adaptive neural network quantization method, called Adaptive Step Size Quantization (ASQ), to resolve this conflict. Specifically, the proposed ASQ method first dynamically adjusts quantization scaling factors through a trained module capable of accommodating different activations. Then, to address the rigid resolution issue inherent in Power of Two (POT) quantization, we propose an efficient non-uniform quantization scheme. We utilize the Power Of Square root of Two (POST) as the basis for exponential quantization, effectively handling the bell-shaped distribution of neural network weights across various bit-widths while maintaining computational efficiency through a Look-Up Table method (LUT). Extensive experimental results demonstrate that the proposed ASQ method is superior to the state-of-the-art QAT approaches. Notably that the ASQ is even competitive compared to full precision baselines, with its 4-bit quantized ResNet34 model improving accuracy by 1.2\% on ImageNet. 
\end{abstract}

%%Graphical abstract
% \begin{graphicalabstract}
% \includegraphics[width=\textwidth]{images/framework.png}
% \end{graphicalabstract}

%%Research highlights
% \begin{highlights}
% \item We propose a novel uniform quantization approach for activations. The proposed ASQ method dynamically adjusts the scaling factors of the activation quantizer through a trained adaptive module, with the aim of minimizing task-specific loss.
% \item For weight quantization, we improved the POT method with rigid resolution issue and proposed POST, a simple yet effective non-uniform quantization method, which achieves more universal and stable quantization performance.
% \item We conduct extensive experiments with various networks for image classification tasks on CIFAR-10 and ImageNet datasets. The results demonstrate that ASQ significantly outperforms the state-of-the-art QAT methods.
% \end{highlights}

%% Keywords
\begin{keyword}
%% keywords here, in the form: keyword \sep keyword
Model compression \sep Quantization aware training \sep Image classification
%% PACS codes here, in the form: \PACS code \sep code

%% MSC codes here, in the form: \MSC code \sep code
%% or \MSC[2008] code \sep code (2000 is the default)

\end{keyword}

\end{frontmatter}

%% Add \usepackage{lineno} before \begin{document} and uncomment 
%% following line to enable line numbers
%% \linenumbers

%% main text
%%

%% Use \section commands to start a section
\section{Introduction}
\label{introduction}
%% Labels are used to cross-reference an item using \ref command.

In recent years, deep neural networks have achieved significant advancements, excelling in domains like computer vision \cite{Zang_Lin_Tang_Wang_Lv_2024, 9449653}, natural language processing \cite{vaswani2017attention}, and speech recognition \cite{peacocke1995introduction}, often surpassing human-level performance. However, as models grow in size and complexity, they pose challenges for deployment on resource-limited edge devices. To address this, researchers have developed model compression techniques that reduce model size and computational demands while preserving performance. Key compression techniques include quantization \cite{PACT,LSQ}, pruning \cite{DBLP:journals/corr/abs-1902-09574, MLSYS2020_6c44dc73}, knowledge distillation \cite{Ahn_2019_CVPR, Yin_2020_CVPR}, low-rank factorization \cite{6638949}, and the development of compact architectures \cite{JMLR:v20:18-598, Howard_2019_ICCV} specifically designed for efficiency.

Quantization stands out as a promising approach to model compression, offering significant advantages in reducing model size and accelerating inference through efficient integer computations \cite{Li_Xu_Lin_Cao_Liu_Sun_Zhang_2024}. These benefits make quantization particularly attractive for deployment on resource-constrained devices. However, as the bit-width used to represent model weights and activations decreases, performance degradation becomes increasingly apparent, especially in extremely low-bit scenarios, e.g., 2-bit or binary (1-bit) representations, where the representational capacity of network is severely limited \cite{gholami2022survey}. Quantization-aware training (QAT) technologies have emerged as an effective quantization method to mitigate performance loss \cite{LSQ, LSQ+}. The core principle of QAT involves training low-bit-width networks using stochastic gradient descent by updating full-precision weights, which are subsequently quantized to lower bit-widths. Therefore, the QAT technologies allow the network to learn to compensate for quantization errors during training, resulting in quantized models that often achieve performance much closer to their full-precision counterparts. %By integrating the quantization process directly into the training pipeline, QAT enables the development of highly efficient, quantized neural networks suitable for a wide range of applications, from mobile devices to large-scale server deployments.

\begin{figure}[t]
    \centering
    \begin{subfigure}[b]{0.45\textwidth}
        \centering
        \includegraphics[width=\textwidth]{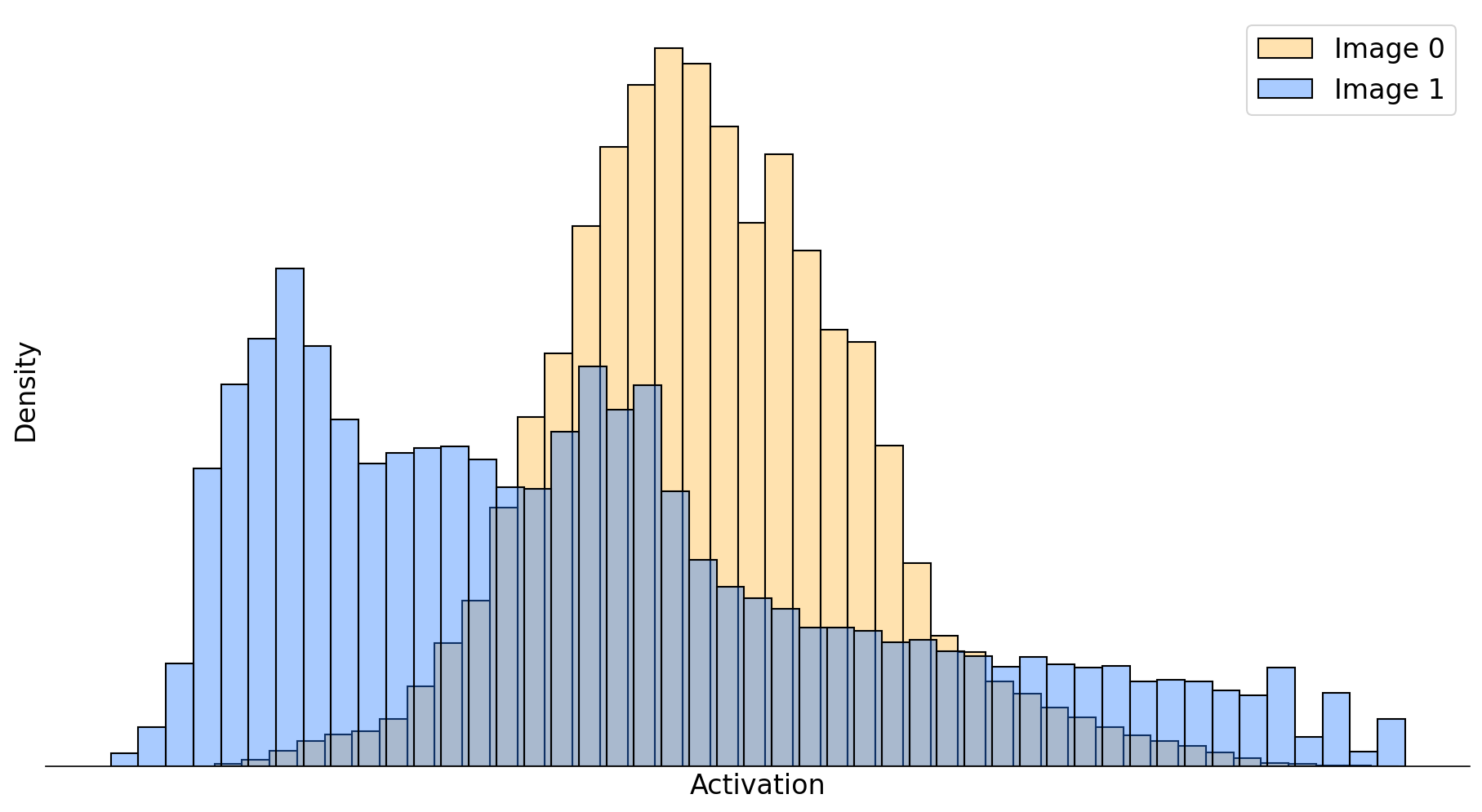}
        % \caption{Distribution of different Activation (inputs) for Deep Neural Network Models.}
        \caption{Activation Distribution.}
        \label{fig:motivation1}
    \end{subfigure}
    \hfill
    \begin{subfigure}[b]{0.45\textwidth}
        \centering
        \includegraphics[width=\textwidth]{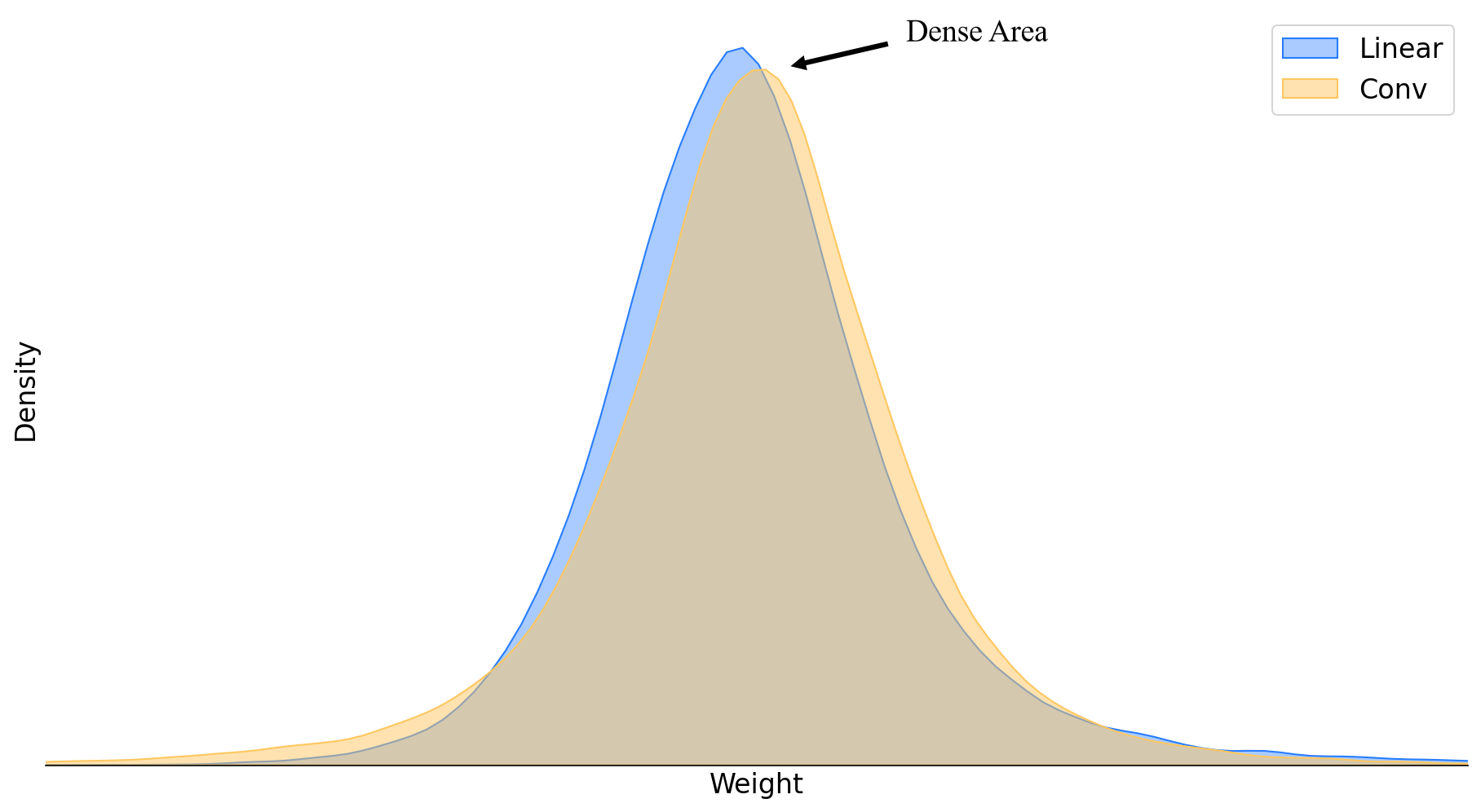}
        % \caption{Weight Distribution in Convolutional Layers and Linear Layers of Deep Neural Network Models.}
        \caption{Weight Distribution.}
        \label{fig:motivation2}
    \end{subfigure}
    
    \caption{(a) The data distributions of different images (model inputs) exhibit significant variations. (b) The distribution of weights in a convolutional layer or linear layer of a deep neural network generally exhibits a bell-shaped curve centered around a mean close to zero.}
    \label{fig:motivation}
\end{figure}

The researches on QAT can be roughly divided into two categories, i.e., gradient estimation and quantization parameter optimization. Among them, the straight-through estimator (STE) is the most mainstream method in the gradient estimation category \cite{STE}. Due to the rounding operation, the quantizer itself is non-differentiable. STE essentially ignores the rounding operation and uses an identity function to approximate the gradient before and after quantization. Despite its simplicity, this approach is often highly effective. In the early stages, quantization parameters were typically determined manually \cite{esser2016cover} or calculated based on the statistical distribution of the data \cite{DBLP}. Although these methods significantly reduced quantization error, they were not optimal for the model's task performance. Later, researchers proposed various methods using backpropagation with stochastic gradient descent to learn quantizer parameters that minimize task loss \cite{PACT,LSQ,LSQ+}. These methods make the optimization of quantizer parameters more direct. Notably, the learned step size quantization (LSQ) method \cite{LSQ} introduced a new gradient estimate to learn the scaling factors (also known as step size) for non-negative activations during QAT. This innovation enabled quantized models to achieve state-of-the-art performance across different bit-widths. However, the method of using trainable quantization parameters has a significant limitation: during the inference phase, each layer of the neural network can only use a fixed quantization step size. This limitation is particularly prominent when dealing with diverse activation value distributions, as illustrated in Fig.\ref{fig:motivation}(a).

Moreover, extensive researches indicate that the weights of deep neural networks typically follow a bell-shaped distribution, as shown in Fig.\ref{fig:motivation}(b), with a significant concentration of values around the mean. Applying uniform quantization to this distribution results in substantial information loss, especially for the critical values near the center. To address these challenges, researchers have proposed various non-uniform quantization methods \cite{QIL,LQ-Nets} that employ different quantization levels to better capture critical features. However, these methods introduce non-linear operators, significantly reducing hardware inference efficiency. Power of two (POT) quantization \cite{Logquant} attempts to balance efficiency and non-uniform distribution by encoding using a logarithmic scale based on powers of two, but its rigid resolution \cite{APOT} prevents it from fully leveraging the benefits of non-uniform distribution.

To better fit the potential distribution of activations and weights, as well as reduce the performance degradation caused by quantization errors, we propose an effective learnable adaptive QAT method, named Adaptive Step size Quantization (ASQ). Based on a dynamic weight technique \cite{Dynamic_Conv}, the proposed ASQ is able to adaptively adjust the model's quantization step size parameters according to the input activation values through a simple neuron module containing one or two linear layers. This module introduces negligible computational and parameter overhead while significantly enhancing the performance of the quantizer. In addition to optimizing activation quantization, we have also improved the POT quantizer for weight quantization. The rigid resolution problem inherent to POT quantization is effectively addressed by using the Power Of Square root of Two (POST) as the exponential quantization basis. This adjustment ensures that even with larger bit widths, the quantization levels do not exhibit excessively high resolution near zero. During actual inference, POST achieves computational efficiency comparable to POT through the use of a look-up table (LUT) method.

Our main contributions are summarized as follows:
\begin{itemize}
    \item We propose a novel uniform quantization approach for activations. The proposed ASQ method dynamically adjusts the step size of the activation quantizer through a trained adaptive module, with the aim of minimizing task-specific loss.
    \item For weight quantization, we improved the POT method with rigid resolution issue and proposed POST, a simple yet effective non-uniform quantization method, which achieves more universal and stable quantization performance.
    \item We conduct extensive experiments with various networks for image classification tasks on CIFAR-10 and ImageNet datasets. The results demonstrate that ASQ significantly outperforms the state-of-the-art QAT methods.
\end{itemize}

%% Use \subsection commands to start a subsection.
\section{Related Works}
\label{related works}

Neural network quantization refers to the process of mapping the weights or activations of a deep neural network to a finite set of values. The quantization techniques can generally be classified into two categories, i.e., \textit{uniform quantization} and \textit{non-uniform quantization}.

\subsection{Uniform Quantization.} 

Uniform quantization is a widely adopted quantization approach. A representative method in this domain is LSQ \cite{LSQ}, which introduces a learnable step size for quantization by improving the configuration of the quantizer. LSQ+ \cite{LSQ+} extends LSQ to address the quantization of negative activation values. DoReFa-Net \cite{DoReFa-Net} proposes training convolutional neural networks with low-bitwidth weights and activations using low-bitwidth parameter gradients. PACT \cite{PACT} employs a novel activation quantization scheme by optimizing the clipping parameters of activations during training, achieving high accuracy even with ultra-low precision weights and activations. DSQ \cite{DSQ} introduces an innovative differentiable soft quantization approach that gradually evolves during training to approximate standard quantization, effectively addressing the discreteness issue in low-bit quantization. 

\subsection{Non-Uniform Quantization.} 

Non-uniform quantization can better capture the distribution characteristics, potentially leading to higher task performance. However, it often requires additional computational overhead, making it challenging to efficiently deploy non-uniform quantization schemes on general-purpose hardware. QIL \cite{QIL} proposes a quantizer that learns the quantization intervals by directly minimizing the task loss of the network to find the optimal quantization intervals. POT \cite{Logquant, INQ} utilizes a logarithmic representation based on powers of two to encode weights and activations. APOT \cite{APOT} introduces an additive POT quantization scheme, which optimizes the overly dense quantization levels in POT by restricting all quantization levels to the sum of powers of two. LQ-Net \cite{LQ-Nets} presents a method for jointly training the quantized network and the quantizer, rather than relying on fixed handcrafted quantization schemes.

%% Use \subsubsection, \paragraph, \subparagraph commands to 
%% start 3rd, 4th and 5th level sections.
%% Refer following link for more details.
%% https://en.wikibooks.org/wiki/LaTeX/Document_Structure#Sectioning_commands

\section{Method}
\label{sec:method}
\begin{figure*}[t]
  \centering
   \includegraphics[width=0.99\textwidth]{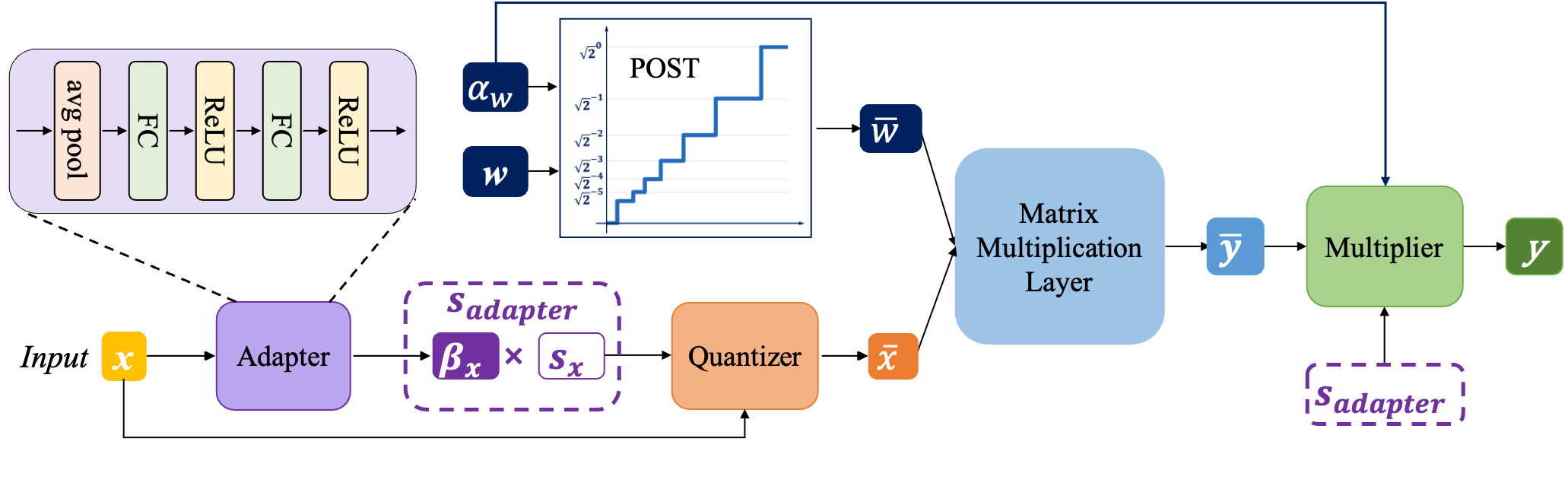}
   \caption{An overview of the proposed method. We optimized quantization for both activations and weights. For activations, a two-layer linear adapter produces an factor $\beta$, which multiplies a trainable parameter $s$ to dynamically adjust the quantization step size. For weights, POST quantization addresses the rigid resolution issue and better fits the bell-shaped distribution.}
   \label{fig:quantizer structure}
\end{figure*}

In this section, we first provide a brief overview of the background of neural network quantization. Subsequently, we discuss the deficiencies of current quantizers and detail our proposed methods. This includes a formal introduction to the uniform quantizer (ASQ) for activations, along with the formulation of related parameter gradients, and an in-depth analysis of the non-uniform quantizer (POST) for weights. Fig.\ref{fig:quantizer structure} is an overview of the proposed method.

\subsection{Preliminaries}
\label{subsec:preliminaries}
To elucidate the concepts, we will use symmetric quantization as our primary example. In uniform quantization, the processes of quantization $Quant$ and dequantization $Dequant$ can be succinctly expressed through the following pair of equations:
\begin{align}
\mathit{Quant:} & \quad x_{int} = clamp(\lfloor \frac{x}{s} \rceil, n, p), \\
\mathit{Dequant:} & \quad \hat{x} = x_{int} \times s,
\end{align}
where $n$ and $p$ refer to the lower and upper bounds of the quantization space, respectively. The operator $\lfloor \cdot \rceil$ represents the rounding operation, while $clamp(\cdot)$ indicates the clipping of values to the range $[n, p]$. The variable $x$ represents the weights or activations undergoing quantization, and $s$ signifies the quantization step size used for value mapping. It's worth noting that the LSQ method poposed in \cite{LSQ} enhances model performance post-quantization by making the step size $s$ a learnable parameter.

Non-uniform quantization can offer additional benefits, but often at the cost of increased computational complexity. POT quantization, however, is an exception. It not only addresses the issue of non-uniform distribution of quantized parameters but also enhances the inference speed in hardware implementations. POT achieves this speed improvement by utilizing bit-shift operations instead of more computationally expensive integer arithmetic. The quantization levels $Q$ in POT can be formally represented by the following equation:
\begin{equation}
Q(\alpha, b) = \alpha \times \{0, \pm2^{-2^{b-1}+1}, \pm2^{-2^{b-1}+2}, ..., \pm1\},
\label{eq:non-uniform_pot}
\end{equation}
where $\alpha$ represents the clipping threshold, and $b$ denotes the bit-width of the fixed-point representation. For both weights and activations, values outside the range $[-\alpha, \alpha]$ are clipped, while those within this range are mapped to the quantization levels defined by $Q(\alpha, b)$. A key characteristic of POT quantization is that it provides higher resolution for values near zero. This property aligns well with the typical distribution of weights in deep neural networks, where smaller values are often more prevalent. %Consequently, in regions where weights are more densely distributed, POT achieves a finer-grained quantization, potentially leading to better preservation of model accuracy.

\subsection{Adaptive Step Size Quantization for Activation}
The LSQ method introduces learnable quantization step size parameters for both weights and activations, enabling the model to adaptively learn appropriate quantization mapping strategies through the stochastic gradient descent algorithm. However, since the activation is scene-sensitive, using a fixed step size to map activations with different distributions to integers may negatively impact the quality of the features, subsequently affecting the overall task performance.

As a solution to the issues discussed above, we propose a general adaptive activation quantization scheme. This approach incorporates a linear layer module within the quantizer to generate dynamic parameter $\beta$. During training, this mechanism not only learns the quantization step size parameter but also dynamically adjusts its magnitude based on varying activation distributions. The corresponding formula is as follows:
% \begin{equation}
%     s_a = s\times \beta
%     \label{eq:adaptive scale}
% \end{equation}
\begin{align}
\mathit{Adapt:} & \quad s_a = s\times \beta,  \\
\mathit{Quant:} & \quad x_{int} = clamp(\lfloor \frac{x}{s_a} \rceil, n, p), \\
\mathit{Dequant:} & \quad \hat{x} = x_{int} \times s,
\end{align}
where $\beta$ is the computed adaptive parameter, while $s_a$ denotes the dynamically adjusted quantization step size.

\subsubsection{Gradient} Following the LSQ method, we compute the gradient of $s_a$ as follows:
\begin{equation}
\frac{\partial \hat{x}}{\partial s_a}=
\begin{cases}
-x/s+\lfloor v/s\rceil, & if\ n<x/s_a<p,\\
n, & if\ x/s_a<n,\\
p, & if\ x/s_a>p.
\end{cases}
\label{eq:gradient of s_a}
\end{equation}

The gradient update of $s$ and $\beta$ is calculated using:
\begin{equation}
    \frac{\partial \hat{x}}{\partial s}=\frac{\partial \hat{x}}{\partial s_a}\frac{\partial s_a}{\partial s}=\frac{\partial \hat{x}}{\partial s}\beta,
    \label{eq:gradient of s}
\end{equation}
\begin{equation}
    \frac{\partial \hat{x}}{\partial \beta}=\frac{\partial \hat{x}}{\partial s_a}\frac{\partial s_a}{\partial \beta}=\frac{\partial \hat{x}}{\partial s}s.
    \label{eq:gradient of beta}
\end{equation}

When calculating the partial derivatives of $s_a$ as show in Eq.(\ref{eq:gradient of s_a}) , STE is utilized for approximating gradient calculations.

\subsubsection{Additional model overhead} The proposed ASQ can effectively improve the performance of the model after quantization. %As mentioned earlier, it introduces an adaptive small module composed of two linear layers. This module also introduces additional parameters and computational overhead to the model. To verify the practicality of our method, we calculated the proportion of additional overhead introduced by the ASQ operator at different bit widths, details can be found in Appendix. For example, at a bit width of 4 bits, the computational load increases by only 0.07\%, and parameter storage space by 5.34\%. Given that quantization significantly reduces overall computational and storage demands, this slight overhead introduced by ASQ remains negligible and entirely acceptable in practical applications.
However, an unavoidable drawback is that this module introduces additional parameters and computational overhead. To verify the practical feasibility of our approach, we calculated the proportion of additional overhead introduced by the ASQ operation at different bit widths.

For a convolutional layer, the number of parameters is determined by the total number of parameters in the convolutional kernels, while for a linear (fully connected) layer, the number of parameters is determined by the weights between the input and output nodes. The formulas for calculating the parameters of convolutional and linear layers are as follows:
\begin{equation}
    Param_{conv}=C_{in}\times C_{out}\times K_w \times K_h,
    \label{eq:param conv}
\end{equation}

\begin{equation}
    Param_{linear}=N_{in}\times N_{out}+N_{out},
    \label{eq:param linear}
\end{equation}
where $C_{in}$ and $C_{out}$ represent the number of input and output channels of the convolutional layer, respectively. $K_w$ and $K_h$ denote the width and height of the convolutional kernel, while $N_{in}$ and $N_{out}$ are the number of input and output nodes of the linear layer, respectively.

The number of operations required for both convolutional and linear layers can be calculated by considering the multiplications performed. For a convolutional layer, the operations involve the number of multiplications performed by each convolutional kernel across the input feature map. Similarly, for a linear layer, the operations involve the number of multiplications and additions between the input and output nodes. The total number of operations for convolutional and linear layers is given by:

\begin{equation}
    OPS_{conv}=C_{in}\times C_{out}\times H_{out} \times W_{out}\times K_w \times K_h,
    \label{eq:ops conv}
\end{equation}

\begin{equation}
    OPS_{linear}=N_{in}\times N_{out},
    \label{eq:ops linear}
\end{equation}
where $H_{out}$ and $W_{out}$ are the output feature map dimensions. Based on the calculated results of the parameters and operations mentioned above, we can further compare modules after specific bit-width quantization. Specifically, during the model quantization process, aside from changes in computational complexity, the storage space for parameters is also significantly reduced. In an unquantized model, each weight is typically stored using 32-bit or 16-bit floating-point numbers, whereas in a quantized model, the storage bit-width for each parameter is greatly reduced. When a model is quantized to a lower bit-width, the bit-width required to store each parameter is reduced to B bits, so the storage space after quantization is:
\begin{equation}
    S_{quantized}=Param \times B \  bits,
    \label{eq:quantized storage}
\end{equation}
where $S_{quantized}$ represents the storage space required for the quantized parameters, and $Param$ denotes the amount of parameters calculated earlier.

After obtaining the number of operations for the model modules through Eq.(\ref{eq:ops conv}) and (\ref{eq:ops linear}), we can further calculate the computational complexity of different quantized models. Typically, the lower the quantization bit-width, the lower the equivalent OPS, which means reduced computational complexity. Assuming the number of operations for the original model is OPS, the quantized model can be compared through the following example:

\begin{equation}
    QOPS_{8-bits} = \frac{OPS}{4},
    \label{eq:quantized ops}
\end{equation}

\begin{equation}
    QOPS_{4-bits} = \frac{OPS}{8},
    \label{eq:quantized ops}
\end{equation}
where $QOPS$ is defined as the reference value for the computational load after quantization, assume its full-precision model is 32 bits.

Based on the above calculations, we can easily determine the additional overhead introduced by our Adapter module. Table 1 presents the data on the proportion of additional parameter storage and computational load introduced by ASQ, using the ResNet18 model as an example, after quantization at different bit-widths.
\begin{table}[h]
\caption{Comparison of the proportion of additional parameters and computational overhead introduced by ASQ at different bit widths. Here, Param(\%) represents the percentage of additional parameters relative to the parameter space of the quantized model, and Compute(\%) represents the percentage of additional computational load relative to the computational load of the quantized model.}
  \centering
  \resizebox{0.6\linewidth}{!}{
  \begin{tabular}{cccc}
  \toprule
  \midrule
    Network&Precision & Param(\%) & Compute(\%) \\
    \midrule
    ResNet18& 8 & 2.67 & 0.04\\
    & 4 & 5.34 & 0.07\\
    & 3 & 7.11 & 0.10\\
    & 2 & 10.67 & 0.14 \\
  \bottomrule
  \end{tabular}}
  \label{tab: overhead}
  \end{table}

% From these data, it can be observed that as the bit width decreases, the storage space and computational load required by the model’s parameters are significantly reduced, leading to an increase in the proportion of storage space and computational overhead introduced by ASQ. However, even at the lowest bit width (2 bits), the increase in computational load is only 0.14\%, and the increase in parameter storage space is 10.67\%. Given that the quantized model itself greatly reduces computational load and parameter storage space, these additional overheads are negligible. Therefore, the slight overhead introduced by ASQ is entirely acceptable in practical applications.

 From these data, it is evident that while our method (ASQ) introduces some additional overhead in terms of parameter storage and computational load, this overhead remains relatively small, even at lower bit-width quantization levels. In practical applications, especially when dealing with activation distributions that are sensitive to specific scenarios, such trade-offs are sometimes necessary to achieve optimal performance. %The slight increase in storage and computational costs introduced by ASQ is minimal compared to the substantial reduction in the model's overall resource requirements due to quantization. 
 Therefore, the additional overhead can be considered acceptable in exchange for the performance benefits it provides, making it a reasonable choice in performance-sensitive applications.

\subsubsection{Training algorithm for ASQ}

When training quantized DNNs using ASQ, we independently apply the ASQ quantizer to the activations of convolutional or fully connected layers. Algorithm 1 summarizes the ASQ training process for the convolutional layer as an example. The complete code and relevant details will be released after the paper is accepted.

\begin{algorithm}[h]
\caption{Training a convolutional layer with ASQ.}
\label{alg:algorithm}
\textbf{Input}: full precision weights of convolutional layer $W_{weight}$ and full precision inputs/activations $A$. \\
\textbf{Parameter}: the quantization step size for weights $\alpha$ and activations $s$, the bit-widths ($b_w$, $b_a$). the weight and bias of adapter $W_{adapter}$ and $B_{adapter}$.\\
\textbf{Output}: updated parameters $W_{weight}$, $W_{adapter}$, $B_{adapter}$, $\alpha$ and $s$.
\begin{algorithmic}[1] %[1] enables line numbers
\STATE Compute the quantized weights using Eq. (1) and Eq. (2): $\hat{W}\gets Quantize(W,\alpha,b_w)$.
\STATE Compute the adaptive factor $\beta$ using Adapter: $\beta \gets Adapter(A,W_{adapter},B_{adapter})$.
\STATE Compute the adaptive quantization step size for activations using Eq. (4): $s_a = s\times \beta$.
\STATE Compute the quantized activations using Eq. (5) and Eq. (6): $\hat{A}\gets Quantize(A,s_a,b_a)$.
\STATE Compute the convolution output: $y \gets \hat{W} \times \hat{A}$.
\STATE Compute the loss $\mathcal{L}$ and the gradients $\frac{\partial \mathcal{L}}{\partial y}$.
\STATE Compute the gradients for the weights $\frac{\partial \mathcal{L}}{\partial y}\frac{\partial y}{\partial W}$.
\STATE Compute the gradients of quantization step size for weights and the adaptive quantization step size for activations: $\frac{\partial \mathcal{L}}{\partial y}\frac{\partial y}{\partial \alpha}$ and $\frac{\partial \mathcal{L}}{\partial y}\frac{\partial y}{\partial s_a}$ based on Eq. (7). 
\STATE Compute the gradients of quantization step size for activations and adaptive factor: $\frac{\partial \mathcal{L}}{\partial y}\frac{\partial y}{\partial s_a}\beta$ and $\frac{\partial \mathcal{L}}{\partial y}\frac{\partial y}{\partial s_a}s$ based on Eq. (8) and Eq. (9).
\STATE Update $W_{weight}$, $W_{adapter}$, $B_{adapter}$, $\alpha$ and $s$ with the corresponding gradients, respectively.
\end{algorithmic}
\end{algorithm}

\subsection{Non-Uniform Quantization for Weight: POST}

For weight quantization, as mentioned in section 3.1, Power of Two (POT) quantization can simultaneously improve performance and efficiency. However, when the bit-width is relatively high, it suffers from a rigid resolution problem. This phenomenon manifests as an uneven distribution of quantization levels, where increasing the bit-width results in excessively fine-grained resolution near zero while leaving large gaps between larger values. Consequently, the model's expressiveness fails to improve effectively, despite the increased bit-width. The method proposed by Additive Power Of Two \cite{APOT} effectively addresses this issue. However, it increases the computational complexity compared to the POT method during the inference process.  In response, we have optimized the POT approach by using the square root of 2 as the base for exponential quantization (Power Of Square root of Two). The quantization levels for this approach is as follows:
\begin{equation}
    Q(\alpha, b)=\\
    \alpha\times\{0, \pm\sqrt{2}^{-2^{b-1}+1}, ..., \pm1\}.
    \label{eq:eq:non-uniform sqrt two}
\end{equation}

\begin{figure*}[t]
    \centering
    \begin{subfigure}[b]{0.31\textwidth}
        \centering
        \includegraphics[width=\textwidth]{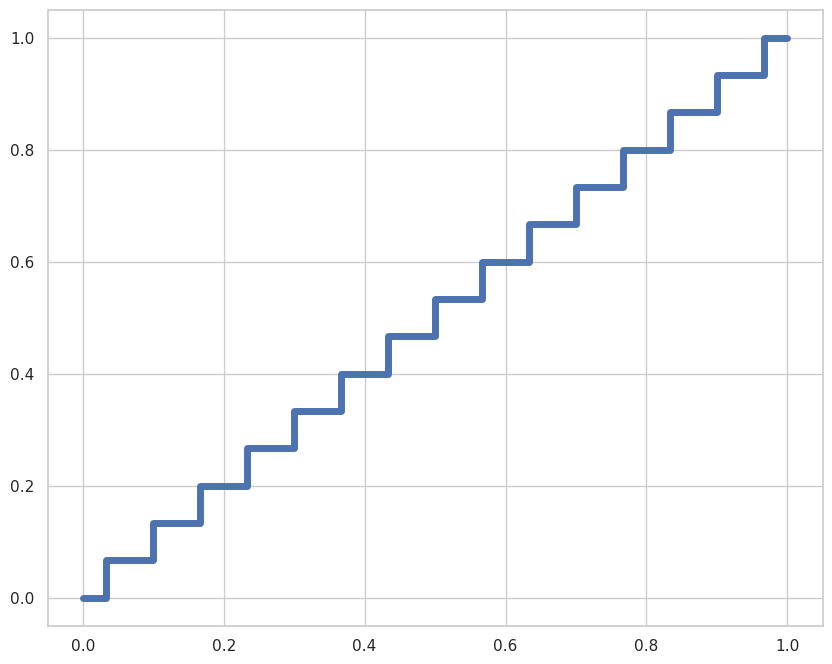} % 替换为你的图片路径
        \caption{Uniform.}
        \label{fig:uniform}
    \end{subfigure}
    \hfill
    \begin{subfigure}[b]{0.31\textwidth}
        \centering
        \includegraphics[width=\textwidth]{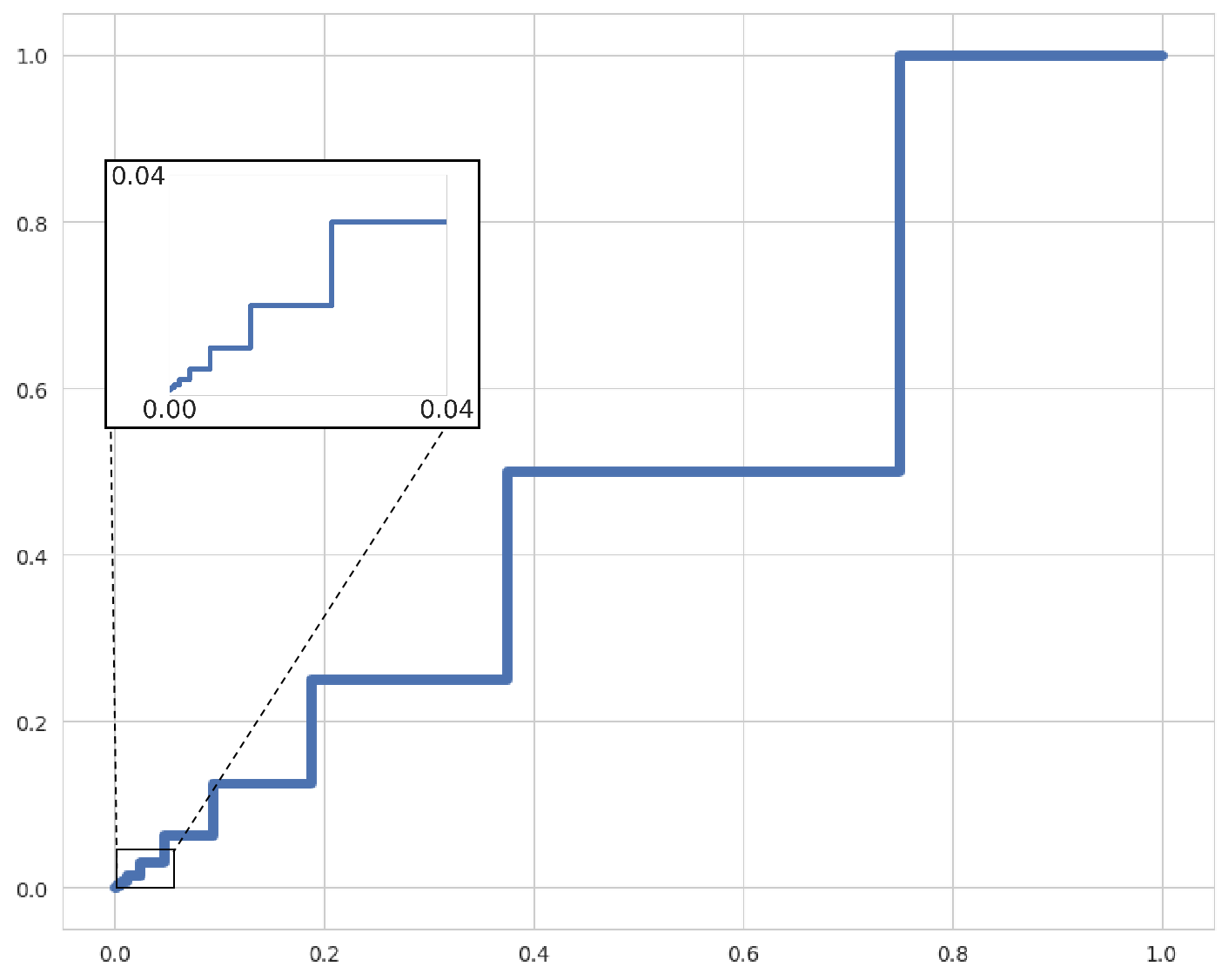} % 替换为你的图片路径
        \caption{Power Of Two.}
        \label{fig:two}
    \end{subfigure}
    \hfill
    \begin{subfigure}[b]{0.31\textwidth}
        \centering
        \includegraphics[width=\textwidth]{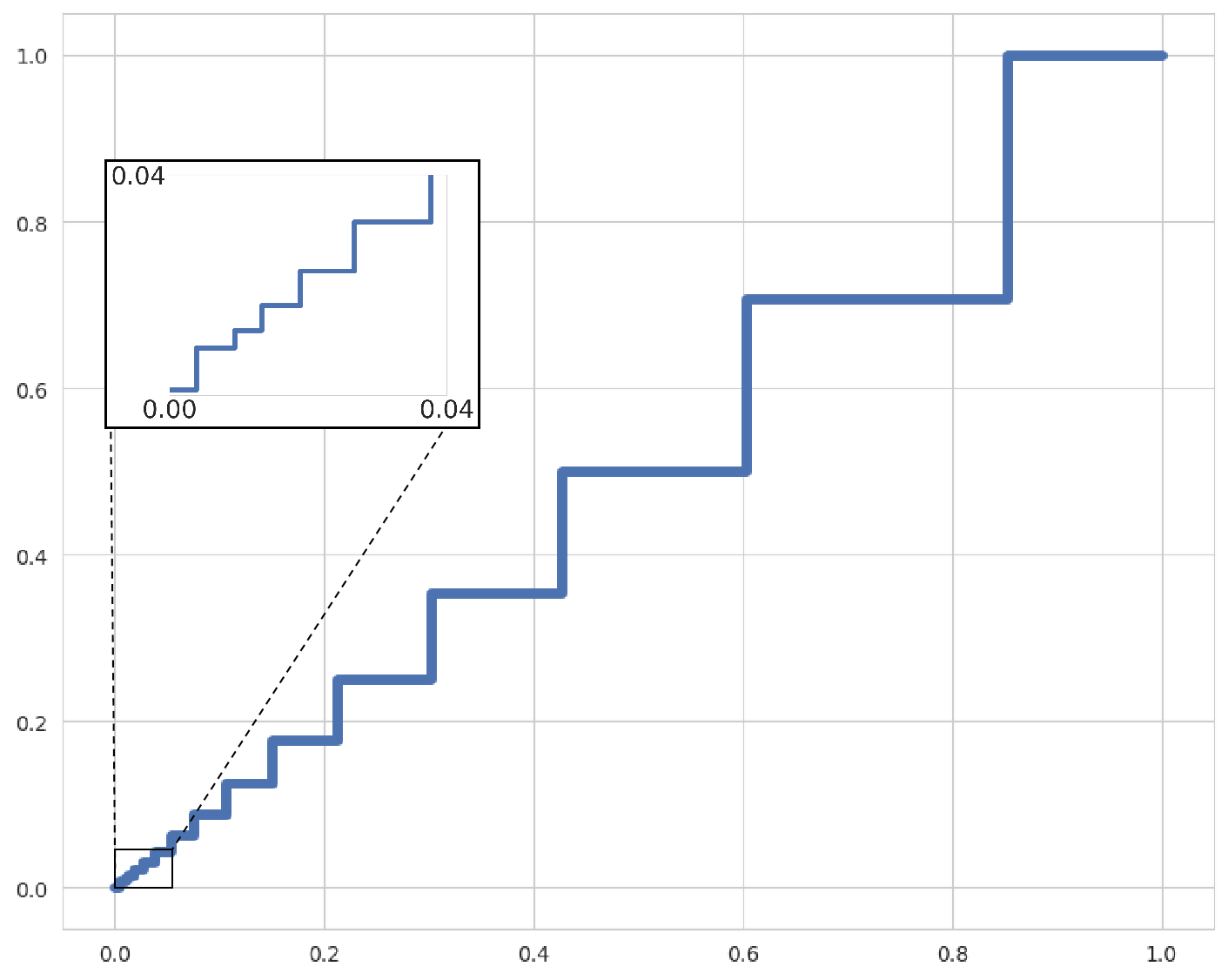} % 替换为你的图片路径
        \caption{Power Of Sqrt of Two.}
        \label{fig:stwo}
    \end{subfigure}
    \caption{The Comparison of 4-bit quantization levels among Uniform, POT, and POST quantization.}
    \label{fig:compare}
\end{figure*}

To better illustrate the effectiveness of POST, we visualized the quantization levels of uniform quantization, POT, and POST under a 4-bit quantization scenario, as shown in Fig.\ref{fig:compare}. By comparing these three figures, we can clearly observe the differences in their quantization effects. Upon careful examination of the magnified portions in Fig.\ref{fig:compare}(b) and Fig.\ref{fig:compare}(c) within the range of 0.00 to 0.04), we can see that the POST method provides coarser quantization levels near zero compared to the POT method. This indicates that the POST scheme effectively mitigates the rigid resolution problem. Simultaneously, we have achieved computational efficiency comparable to POT (Power Of Two) by employing a Look-Up Table (LUT) method, with only a slight increase in space utilization.

\subsubsection{Computation.} In hardware computation, multiplying an integer power of 2, such as $2^n$, by another integer $r$ can be efficiently achieved through simple bit-shifting, bypassing the need for a complex multiplier. A similar approach can be applied to the square root of two. Specifically, we can equate $\sqrt{2}^n$ to $2^{(\frac{n}{2})}$. When n is even, $\frac{n}{2}$ is an integer, allowing for simple bitwise shifting of r to perform the multiplication. However, when n is odd, $\frac{n}{2}$ is not an integer, rendering the shifting technique ineffective. For low-bit scenarios, where the integer space is limited, a Look-Up Table (LUT) method can effectively address this special case. For example, when quantizing a model to 3 bits, a table of size [8,4] is sufficient to handle all cases where n is odd in integer operations within neural networks. This computation method can be represented by the following expression:

\begin{equation}
    \sqrt{2}^{-\mathbf{W}}\mathbf{A} =
    \begin{cases}
    \mathbf{A}, & if\ \mathbf{W}=0,\\
    \mathbf{A} >>\mathbf{W}/2, & if\ \mathbf{W}\ is\ even,\\
    LUT(\mathbf{W}, \mathbf{A}), & if\ \mathbf{W}\ is\ odd.
    \end{cases}
    \label{eq:inference}
\end{equation}
Here, $\mathbf{W}$ and $\mathbf{A}$ represent the weights and activations in the deep neural network that have been mapped to fixed-point numbers with a specific bit-width. Compared to uniform quantization, our method achieves approximately $b$-fold faster multiplication operations, where $b$ denotes the bit width used in the quantization process.

\section{Experiments}
\label{sec:experiments}
In this section, we validate our proposed method on vanilla ResNet \cite{ResNet} and MobileNet-V2 \cite{MobileNetV2} using ImageNet-ILSVRC2012 \cite{ImageNet} and CIFAR10 \cite{CIFAR} datasets. We also conduct an ablation study on the various components of our algorithm. All experiments were implemented using PyTorch.

\subsection{Implimention Details}
\subsubsection{Basic setup.} We established two quantization schemes: Scheme 1 employs pure uniform quantization, using our proposed ASQ (Adaptive Step size Quantization) for activations and a fixed trainable step size uniform quantization scheme (LSQ - Learned Step Size Quantization) for weights. Scheme 2 adopts a non-uniform quantization approach for weights: while still using ASQ for activations, it employs a POST (Power Of Square of Two) quantizer for non-uniform exponential mapping of weights. For Scheme 1, we conducted comparisons across 2, 3, 4, and 8-bit quantization. For Scheme 2, we only compared 3 and 4-bit widths. This is because 8-bit quantization results in negligible information loss, and using POST in actual inference would introduce significant additional space overhead. With 2-bit quantization, applying non-uniform symmetric quantization to weights would actually turn it into ternarisation, negating the benefits of non-uniform quantization.

\subsubsection{Training.} Following the training scheme outlined in LSQ \cite{LSQ}, We utilized the SGD optimizer and a cosine learning rate decay without restarts \cite{Sgdr}. All weights in the quantized model were initialized using the official PyTorch pre-trained full-precision model. The Adapter will be initialized to produce output values close to 1. This initialization strategy is designed to ensure that the adapter does not make excessive adjustments to the quantization step size in the initial stages. %For the training hyperparameters, the batch size was set to 512. and We applied a weight decay of 1e-4. Basic data augmentation was used, similar to ResNet\cite{ResNet}. Training images were randomly resized and cropped to 224 × 224 pixels and randomly horizontally flipped. Testing images were center-cropped to 224 × 224 pixels.

\subsubsection{Architecture.} For a fair comparison, we employed vanilla ResNet and MobileNet-V2 architectures without any modifications. All convolutional and linear layers were quantized to low bit-widths, except for the first and last layers of the model, which were consistently quantized to 8-bit.

\subsection{Evaluation on ImageNet}

We evaluated the performance of ASQ and POST on the ResNet18/34 and MobileNet-V2 models using the ImageNet dataset. The models were trained for 90 epochs with an initial learning rate of 0.01 for the weights. The batch size was set to 512 for ResNet-18/34 and 256 for MobileNet-V2. Additionally, the weight decay was set to 1e-4. The training images were resized, cropped to 224 × 224 pixels, and randomly flipped horizontally. The test images were center-cropped to 224 × 224 pixels.

\begin{table*}[h]
\caption{Comparison of low precision Resnet on ImageNet2012. Methods include PACT \cite{PACT}, LQ-Net \cite{LQ-Nets}, DSQ \cite{DSQ}, FAQ \cite{FAQ}, QIL \cite{QIL}, APOT \cite{APOT}, LSQ \cite{LSQ}, DAQ \cite{DAQ}, LIMPQ \cite{LIMPQ}, SEAM \cite{SEAM}, RMQ \cite{RMQ}. $\dagger$ indicates that the method uses mixed-precision quantization. $*$ our re-implementation. To ensure a fair comparison, we have listed the accuracy of the full-precision baseline model used by all the methods we compare. All subsequent comparison tables follow the same approach.}
  \centering
  \resizebox{\linewidth}{!}{
  \begin{tabular}{cc|ccccc|ccccc}
  \toprule
   \multirow{2}{*}{Network} & \multirow{2}{*}{Method} & \multicolumn{5}{c|}{Top-1 Accuracy @ Precision} & \multicolumn{5}{c}{Top-5 Accuracy @ Precision} \\
    % \cmidrule(r){3-6} \cmidrule(r){7-10}
    & & 32 & 2 & 3 & 4 & 8 & 32 & 2 & 3 & 4 & 8 \\
    \midrule
    \multirow{13}{*}{ResNet18}& PACT & 70.4 & 64.4(-6.0) & 68.1(-2.3) & 69.2(-1.2) & - & 89.6 & 85.6(-4.0) & 88.2(-1.4) & 89.0(-0.6) & - \\
    & LQ-Net & 70.3 & 64.9(-5.4) & 68.2(-2.1) & 69.3(-1.0) & - & 89.5 & 84.3(-5.2) & 85.9(-3.6) & 88.8(-0.7) & - \\
    & DSQ & 69.9 & 65.2(-4.7) & 68.7(-1.2) & 69.6(-0.3) & - & - & - & - & - & - \\
    & FAQ & 69.8 & - & - & 69.8(+0.0) & 70.0(+0.2) & 89.1 & - & - & 89.1(+0.0) & 89.3(+0.2) \\
    & QIL & 70.2 & 65.7(-4.5) & 69.2(-1.0) & 70.1(-0.1) & - & - & - & - & - & - \\
    % \rowcolor{red} & EWGS\cite{EWGS} & 69.9 & 67.0(-2.9) & 69.7(-0.2) & 70.6(+0.7) & - & - & - & - & - & - \\
    & APOT & 69.8 & 66.5(-3.3) & 69.8(+0.0) & 70.7(+0.9) & - & 89.4 & 87.5(-1.9) & 89.2(-0.2) & 89.6(+0.2) & - \\
    & $\text{LSQ}^*$ & 69.8 & 66.3(-3.5) & 69.3(-0.5) & 70.3(+0.5) & 70.7(+0.9) & 89.1 & 87.0(-2.1) & 89.0(-0.1) & 89.5(+0.4) & 89.7(+0.6) \\
     & DAQ & 69.9 & 66.9(-3.0) & 69.6(-0.3) & 70.5(+0.6) & - & - & - & - & - & - \\
    & $\text{LIMPQ}^\dagger$ & 70.5 & - & 69.7(-0.8) & 70.8(+0.3) & - & - & - & - & - & - \\
    & $\text{SEAM}^\dagger$ & 70.5 & - & 70.0(-0.5) & 70.8(+0.3) & - & - & - & - & - & - \\
    & $\text{RMQ}^\dagger$ & 70.5 & - & 70.2(-0.3) & 71.0(+0.5) & - & - & - & - & - & - \\
    % \rowcolor[HTML]{fffbcc} & ASQ(ours) & 69.8 & \textbf{67.1(-2.7)} & 69.8(+0.0) & 70.8(+1.0) & \textbf{71.0(+1.2)} & 89.1 & \textbf{87.4(-1.7)} & 89.2(+0.1) & 89.7(+0.6) & \textbf{89.8(+0.7)} \\
    % \rowcolor[HTML]{fffbcc} & ASQ+POST(ours) & 69.8 & - & \textbf{70.0(+0.2)} & \textbf{71.0(+1.2)} & - & 89.1 & - & \textbf{89.3(+0.2)} & \textbf{89.9(+0.8)} & - \\
    & ASQ(ours) & 69.8 & \textbf{67.1(-2.7)} & 69.8(+0.0) & 70.8(+1.0) & \textbf{71.0(+1.2)} & 89.1 & \textbf{87.4(-1.7)} & 89.2(+0.1) & 89.7(+0.6) & \textbf{89.8(+0.7)} \\
    & ASQ+POST(ours) & 69.8 & - & \textbf{70.0(+0.2)} & \textbf{71.0(+1.2)} & - & 89.1 & - & \textbf{89.3(+0.2)} & \textbf{89.9(+0.8)} & - \\
  \midrule
   \multirow{9}{*}{ResNet34}& LQ-Net & 73.8 & 69.8(-4.0) & 71.9(-1.9) & - & - & 91.4 & 89.1(-2.3) & 90.2(-1.1) & - & - \\
   & DSQ & 73.8 & 70.0(-3.8) & 72.5(-1.3) & 72.8(-1.0) & - & - & - & - & - & - \\
   & FAQ & 73.3 & - & - & 73.3(+0.0) & 73.7(+0.2) & 91.4 & - & - & 91.3(-0.1) & 91.6(+0.2) \\
   & QIL & 73.7 & 70.6(-3.1) & 73.1(-0.6) & 73.7(+0.0) & - & - & - & - & - & - \\
   % \rowcolor{red} & EWGS\cite{EWGS} & 73.3 & 71.4(-1.9) & 73.3(+0.0) & 73.9(+0.6) & - & - & - & - & - & - \\
   & APOT & 73.7 & 70.9(-2.8) & 73.4(-0.3) & 73.8(+0.1) & - & 91.3 & 89.7(-1.6) & 91.1(-0.2) & \textbf{91.6(+0.3)} & - \\
   & $\text{LSQ}^*$ & 73.3 & 70.6(-2.7) & 73.1(-0.2) & 73.7(+0.4) & 74.1(+0.8) & 91.4 & 89.8(-1.6) & 91.3(-0.1) & 91.5(+0.1) & \textbf{91.8(+0.4)} \\
   & DAQ & 73.3 & 71.0(-2.3) & 73.1(-0.2) & 73.7(+0.4) & - & - & - & - & - & - \\
   % \rowcolor[HTML]{fffbcc} & ASQ(ours) & 73.3 & \textbf{71.3(-2.0)} & 73.3(+0.0) & 73.9(+0.6) & \textbf{74.3(+1.0)} & 91.4 & \textbf{90.2(-1.2)} & \textbf{91.4(+0.0)} & 91.6(+0.2) & \textbf{91.8(+0.4)} \\
   % \rowcolor[HTML]{fffbcc} & ASQ+POST(ours) & 73.3 & - & \textbf{73.4(+0.1)}& \textbf{74.1(+0.8)} & - & 91.4 & - & \textbf{91.4(+0.0)} & \textbf{91.7(+0.3)} & - \\
   & ASQ(ours) & 73.3 & \textbf{71.3(-2.0)} & 73.3(+0.0) & 73.9(+0.6) & \textbf{74.3(+1.0)} & 91.4 & \textbf{90.2(-1.2)} & \textbf{91.4(+0.0)} & 91.6(+0.2) & \textbf{91.8(+0.4)} \\
   & ASQ+POST(ours) & 73.3 & - & \textbf{73.4(+0.1)}& \textbf{74.1(+0.8)} & - & 91.4 & - & \textbf{91.4(+0.0)} & \textbf{91.7(+0.3)} & - \\
  \bottomrule
  \end{tabular}}
  \label{tab: sota0}
  \end{table*}

Table \ref{tab: sota0} presents a comparison of the accuracy of two quantization strategies (ASQ and ASQ+POST) on ResNet models against other state-of-the-art (SOTA) models. Whether compared to uniform or non-uniform quantization, the results demonstrate a significant performance advantage for both strategies. It is worth noting that non-uniform quantization strategies show more pronounced effects at larger bit widths, as the larger discrete value set provides more flexible training space for logarithmic quantization. It can be observed that our 4-bit and 8-bit quantized networks achieve higher accuracy than the full-precision baseline network (On ResNet18, ASQ achieved a 1.0\% accuracy improvement with 4-bit quantization and a 1.2\% improvement with 8-bit quantization. With ASQ+POST, the 4-bit quantization improved accuracy by 1.2\%. On ResNet34, ASQ improved accuracy by 0.6\% with 4-bit quantization and 1.0\% with 8-bit quantization, while ASQ+POST achieved a 0.8\% improvement with 4-bit quantization.).

The 3-bit quantized network with ASQ also maintained the accuracy of the full-precision baseline model. When the bit-width was further reduced to 2, although there was a more significant drop in accuracy compared to the baseline model, our approach still outperformed other methods. These results indicate that dynamically adjusting the quantization step size of activations by training the adaptive module (Adapter), along with using a non-uniform quantization scheme, can indeed effectively improve the accuracy of quantized models. Moreover, when using our quantization scheme 2 (which uses POST quantization for weights), our model achieves better hardware performance in terms of inference speed. 

\begin{table}[h]
\caption{Comparison of low precision MobileNet-V2 on ImageNet2012. Methods include PACT \cite{PACT}, DSQ \cite{DSQ}, NIPQ \cite{NIPQ}, LSQ \cite{LSQ}. $\dagger$ indicates that the method uses mixed-precision quantization. $*$ our re-implementation.}
  \centering
  \resizebox{0.65\linewidth}{!}{
  \begin{tabular}{cc|ccc}
  \toprule
   \multirow{2}{*}{Network} & \multirow{2}{*}{Method} & \multicolumn{3}{c}{Top-1 Accuracy @ Precision} \\
    % \cmidrule(r){3-6} \cmidrule(r){7-10}
    & & 32 & 3 & 4 \\
    \midrule
    \multirow{6}{*}{MobileNet-V2}& PACT & 71.8 & - & 61.4(-10.4)\\
    & DSQ & 71.9 & - & 64.8(-7.1) \\
    & $\text{NIPQ}^\dagger$ & 72.6 & 62.3(-10.3) & 69.2(-3.4) \\
    & $\text{LSQ}^*$ & 71.9 & 65.2(-6.7) & 69.5(-2.4) \\
    & ASQ(ours) & 71.9 & 65.6(-6.3) & 69.8(-2.1) \\
    & ASQ+POST(ours) & 71.9 & \textbf{66.6(-5.3)} & \textbf{70.1(-1.8)} \\
  \bottomrule
  \end{tabular}}
  
  \label{tab: sota1}
\end{table}

Even on the compact and efficient MobileNet-V2 model, our method demonstrates improvements compared to other approaches, as shown in Table \ref{tab: sota1}. Although the gains are not as pronounced as those observed on ResNet models, this still highlights the generalizability of our approach.

\subsection{Evaluation on CIFAR-10}
We also conducted experiments on the CIFAR-10 dataset. For these experiments, we set the initial learning rate for the weights to 0.04, with a batch size of 128. The weight decay was set to 1e-4, and momentum was set to 0.9. Each quantized model was trained for 300 epochs. For the CIFAR-10 dataset, we applied standard data augmentation techniques, including random cropping and horizontal flipping.

\begin{table}[h]
\caption{Comparison of low precision ResNet on CIFAR-10. Methods include LQ-Net \cite{LQ-Nets}, LSQ \cite{LSQ}, APOT \cite{APOT}. $*$ our re-implementation.}
  \centering
  \resizebox{0.75\linewidth}{!}{
  \begin{tabular}{cc|cccc}
  \toprule
   \multirow{2}{*}{Network} & \multirow{2}{*}{Method} & \multicolumn{4}{c}{Top-1 Accuracy @ Precision} \\
    % \cmidrule(r){3-6} \cmidrule(r){7-10}
    & & 32 & 2 & 3 & 4 \\
    \midrule
    \multirow{5}{*}{ResNet20}& LQ-Net & 92.1 & 90.2(-1.9) & 91.6(-0.5) & -\\
    & $\text{LSQ}^*$ & 92.6 & 91.0(-1.6) & 92.0(-0.6) & 92.4(-0.2) \\
    & $\text{APOT}^*$ & 92.6 & 91.0(-1.6) & 92.2(-0.4) & 92.7(+0.1) \\
    & ASQ(ours) & 92.6 & \textbf{91.2(-1.4)} & 91.7(-0.9) & 92.8(+0.2)\\
    & ASQ+POST(ours) & 92.6 & - & \textbf{92.3(-0.3)} & \textbf{93.0(+0.4)}\\
    % \midrule
    % \multirow{2}{*}{ResNet56}& LSQ^* & 94.3 & - & - & 94.3(-1.0) \\
    % & APOT^* & - & - & - & - \\
    % \rowcolor[HTML]{fffbcc} & ASQ(ours) & 94.3 & - & - & 94.8(-0.5)\\
  \bottomrule
  \end{tabular}}
  \label{tab: sota2}
  \end{table}

Table \ref{tab: sota2} summarizes the accuracy comparison between our method and other approaches against the baseline model. For the 3-bit and 4-bit models, both quantization schemes achieved results comparable to the full-precision baseline model. Notably, our method outperformed both the uniform quantization method LSQ \cite{LSQ} and the non-uniform quantization methods LQ-Nets \cite{LQ-Nets} and APoT \cite{APOT} across all bit widths from 2 to 4.

\subsection{Ablation Studies}
The proposed algorithm incorporates two key techniques: an Adaptive Step Size Quantization (ASQ) for activation quantization and a Power Of Square root of Two (POST) quantization for weight quantization. To assess the impact of each component on overall performance, we conducted a thorough ablation study on the 4-bit quantized ResNet20 network using the CIFAR-10 dataset.

\begin{table}[h]
\caption{Effects of different components of our method on the final performance of a 3-bit quantized ResNet-20 network. Baseline refers to the quantization strategy where both activations and weights are quantized using the learnable step size quantization \cite{LSQ} method.}
  \centering
  \resizebox{0.6\linewidth}{!}{
  \begin{tabular}{cc}
  \toprule
  \midrule
   {Method} & Top-1 Acc \\
    \midrule
    Baseline(our implementation) & 92.4\\
    Baseline+POT(for weight) & 92.5 \\
    Baseline+POST(for weight) & 92.7 \\
    Baseline+ASQ(for activation) & 92.8 \\
    Baseline+POST(for weight)+ASQ(for activation) & 93.0 \\
    \midrule
    Corresponding real-valued network & 92.6 \\
  \bottomrule
  \end{tabular}}
  \label{tab: ablation}
  \end{table}
As shown in Table \ref{tab: ablation}, the Baseline refers to the method where both activations and weights are quantized using a trainable step size quantizer (LSQ). It can be observed that employing the POT quantizer for weights does improve quantization performance. However, the improvement is limited due to the rigid resolution issue mentioned earlier. This limitation can be effectively addressed by replacing the POT quantizer with the POST quantizer. Furthermore, optimizing the quantization of activations with the ASQ alone also significantly enhances the accuracy of the quantized model. When both the ASQ is applied to activations and the POST quantizer is used for weights, the quantized model's accuracy improves by 0.4\% compared to the real-valued model.

\begin{figure*}[h]
  \centering
   \includegraphics[width=\textwidth]{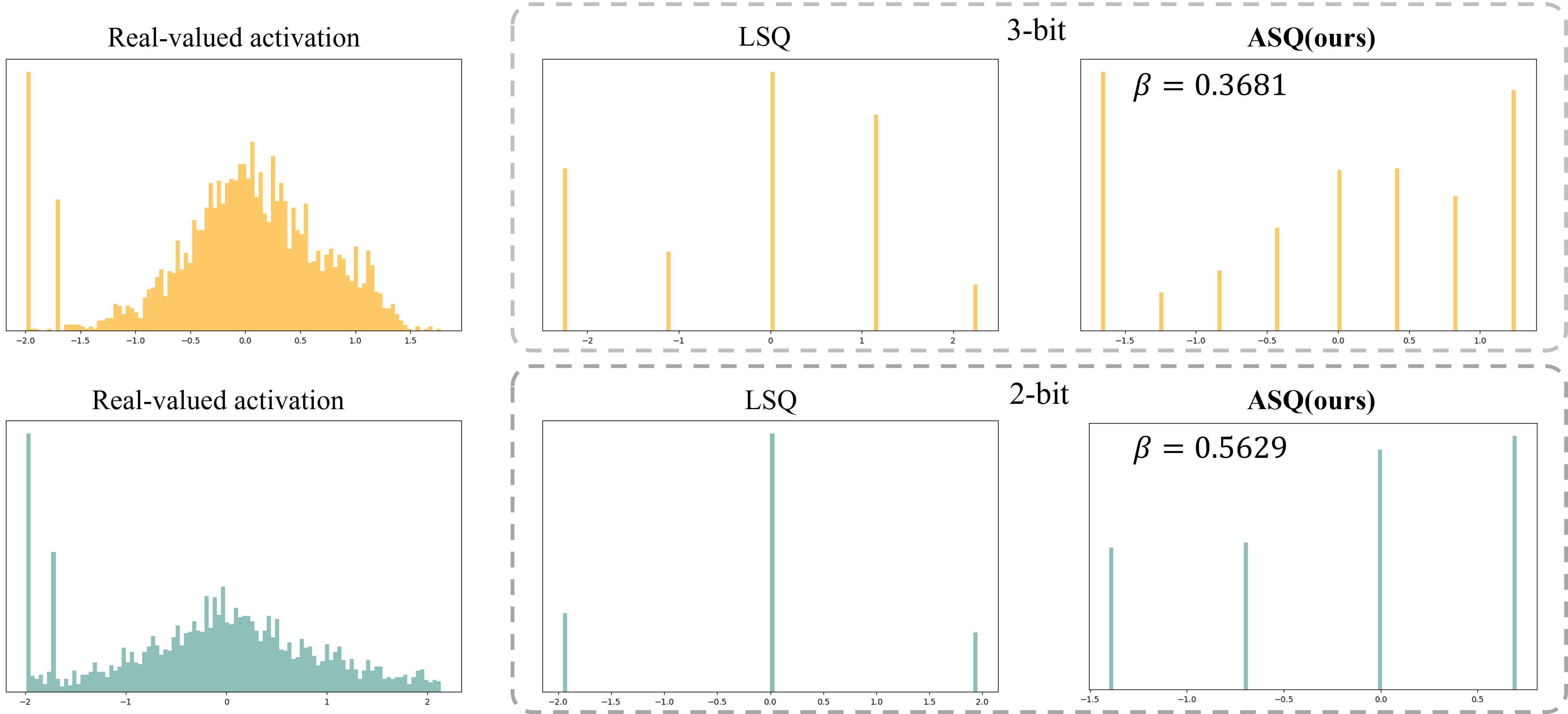}
   \caption{Compare the histograms of the activation distribution before and after quantization using ASQ and LSQ for 2-bit and 3-bit quantization. $\beta$ is the output of the adaptive module used to adjust the quantization step size.}
   \label{fig:visualization}
\end{figure*}

\subsection{Visualization}

% To further understand the proposed ASQ and visually demonstrate its advantages, we visualized the distribution of activations before and after quantization, comparing them with LSQ, as shown in Fig.\ref{fig:visualization}. During inference, when a fixed quantization step size is applied to all activations, there can be an issue of underutilization of the quantized integer space, particularly if the quantization step size is too large for certain activations. For example, when quantizing activation values to 2 bits, the integer space can theoretically represent four distinct values. However, with LSQ, the quantized activations may end up occupying only three discrete values. This underutilization results in increased information loss, which can, in turn, negatively impact the model's performance on the task at hand. ASQ addresses this issue by incorporating an adaptive module that dynamically adjusts the quantization step size to better fit the varying distribution of activations. This adaptability allows ASQ to more effectively utilize the available integer space, reducing the risk of information loss and thereby enhancing the overall performance of the quantized model. 
To better understand the proposed ASQ (Adaptive Step size Quantization) and its benefits, we visualized the distribution of activations before and after quantization, comparing them with LSQ (Learned Step Quantization), as shown in Fig.\ref{fig:visualization}. Fixed quantization step sizes during inference can lead to underutilization of the quantized integer space, especially when the step size is too large for certain activations. For instance, in 2-bit quantization, the integer space can represent four values, but LSQ may ends up using only three, leading to information loss and degraded model performance. ASQ addresses this by dynamically adjusting the quantization step size to match the varying activation distributions, better utilizing the integer space and minimizing information loss, thereby improving model performance.

% The advantages of ASQ extend even further. As depicted in Fig.\ref{fig:block_errors}, we conducted a detailed analysis comparing the impact of LSQ and ASQ quantizers on the outputs of blocks 2 through 9 in ResNet20. This was done by calculating the L2 norm between the outputs generated with full-precision activations and those produced with 3-bit quantized activations. The resulting visualization provides a clear depiction of the errors introduced by each block following quantization, as well as the cumulative trend of error accumulation across the network.  This comparison underscores ASQ's ability to effectively mitigate the quantization-induced discrepancies at each stage, thereby reducing the compounding of errors as the data progresses through the network. The distinct superiority of ASQ over LSQ in this context further highlights its potential in preserving model accuracy, even when subjected to aggressive quantization.

\begin{figure}[H]
  \centering
   \includegraphics[width=0.8\textwidth]{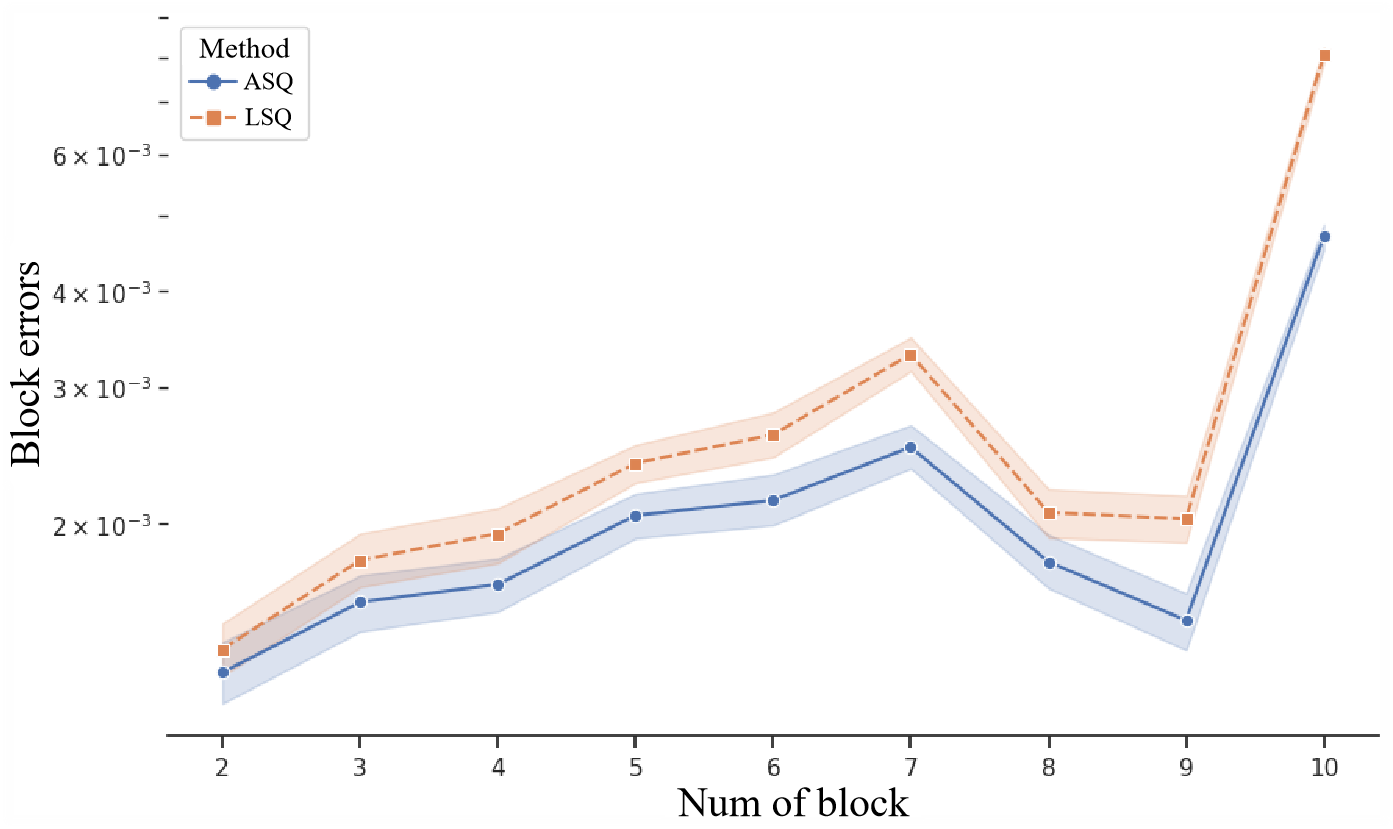}
   \caption{The output error (error accumulation) in blocks 2-9 of ResNet20 using ASQ and LSQ quantizers respectively.}
   \label{fig:block_errors}
\end{figure}

Additionally, as illustrated in Fig.\ref{fig:block_errors}, we compared the impact of LSQ and ASQ on the outputs of blocks 2 through 9 in ResNet20 by calculating the L2 norm between full-precision and 3-bit quantized activations. This analysis highlights the errors introduced by each block and the cumulative error trend across the network. The results demonstrate ASQ's effectiveness in reducing quantization-induced errors at each stage, mitigating error accumulation, and preserving model accuracy even under aggressive quantization, underscoring its superiority over LSQ.

\begin{figure}[h]
    \centering
    \begin{subfigure}[b]{0.7\textwidth}
        \centering
        \includegraphics[width=\textwidth]{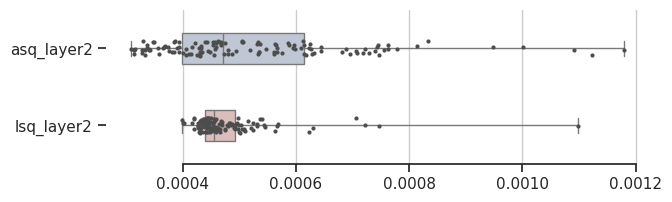} % 替换为你的图片路径
        %\caption{Power Of Two.}
        \label{fig:layer2}
    \end{subfigure}
    \hspace{0pt}
    \begin{subfigure}[b]{0.7\textwidth}
        \centering
        \includegraphics[width=\textwidth]{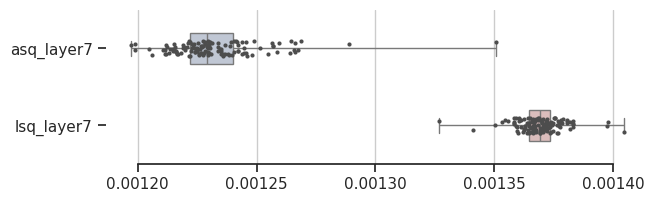} % 替换为你的图片路径
        %\caption{Power Of Sqrt of Two.}
        \label{fig:layer7}
    \end{subfigure}
    \hspace{0pt}
    \begin{subfigure}[b]{0.7\textwidth}
        \centering
        \includegraphics[width=\textwidth]{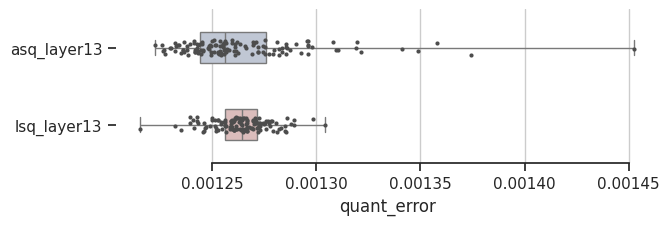} % 替换为你的图片路径
        %\caption{Power Of Sqrt of Two.}
        \label{fig:layer13}
    \end{subfigure}
    \caption{Quantization error of ASQ and LSQ quantizers for Activations in the 2nd, 7th, and 13th layers of ResNet20.}
    \label{fig:quant errors}
\end{figure}

Fig.\ref{fig:quant errors} presents a detailed examination of the quantization error encountered during actual inference in a ResNet20 model, specifically due to the rounding and clipping operations within LSQ and ASQ quantization techniques. The error is quantified as the L2 norm, measuring the deviation between activation values before and after the quantization process.

An intriguing observation emerges from these results: the adaptive parameters introduced by ASQ, while designed to optimize quantization, do not universally result in a reduction of quantization error. In fact, in several instances, the quantization error under ASQ surpasses that of the fixed-step size quantizer, LSQ. This is particularly notable in certain layers where ASQ's error not only exceeds LSQ's but does so with a significantly higher variance.

This behavior underscores a crucial insight: a smaller quantization error does not inherently guarantee enhanced model performance. The adaptive nature of ASQ provides it with a unique ability to adjust dynamically, potentially compensating for the inherent information loss that quantization introduces. Consequently, even in scenarios where ASQ exhibits a larger or more variable quantization error, it can still maintain, or even improve, the model's overall accuracy. This adaptability is particularly beneficial in complex models, where maintaining a balance between quantization error and model performance is critical.

These findings suggest that the effectiveness of a quantization method cannot be solely judged by the magnitude of the quantization error it produces. Instead, one must consider how well the method can preserve the integrity of the model's performance amidst these quantization challenges.

\section{Conclusion}
\label{sec:conclusion}
% In this paper, we propose two novel methods to optimize low-bit-width quantization for improved accuracy: Adaptive Step Size Quantization (ASQ) and Power of Two Square Root Quantization (POST). Based on our observation that low-bit quantizers are highly sensitive to changes in activation distribution, we introduce ASQ, which employs a dynamic weight design to create adaptive quantization step sizes for activations. This approach effectively adjusts the quantizer's step size when faced with different input distributions. Additionally, we develop the POST quantization technique to address the inherent rigid resolution issues of Power-of-Two (POT) quantization while achieving comparable inference speed, with minimal additional memory overhead. We conducted extensive experiments on large and small datasets for image classification tasks using low-bit models, and the results demonstrate that ASQ and POST significantly enhance the performance of quantized models, with accuracy surpassing or matching that of traditional uniform and non-uniform quantization techniques. Furthermore, our ablation studies and visual analysis provide strong evidence that ASQ and POST effectively mitigate and compensate for information loss due to quantization errors. These results not only highlight the potential of our methods but also underscore the advantages of our techniques.

In this paper, we propose a novel methods, called ASQ, to optimize low-bit-width quantization for improved accuracy. Recognizing the sensitivity of low-bit quantizers to activation distribution changes, the poposed ASQ method introduces a dynamic weight design to create adaptive quantization step sizes, effectively adjusting to varying input distributions. The POST for weight introduced in ASQ addresses the rigid resolution issues of POT quantization while maintaining comparable inference speed with minimal memory overhead. Extensive experiments on image classification tasks show that the proposed ASQ significantly enhances quantized model performance, surpassing or matching traditional quantization methods. Ablation studies and visual analyses further confirm that ASQ effectively mitigates information loss due to quantization errors, underscoring the advantages of our techniques.

\section{Acknowledgement}

This work is supported by the Natural Science Foundation of Sichuan under Grant 24NSFSC3404 and 2023NSFSC0474, the National Major Scientific Instruments and Equipments Development Project of National Natural Science Foundation of China under Grant 62427820,  the Fundamental Research Funds for the Central Universities under Grant 1082204112364, the Tianfu Yongxing Laboratory Organized Research Project Funding under Grant 2023CXXM14, and the Science Fund for Creative Research Groups of Sichuan Province Natural Science Foundation under Grant 2024NSFTD0035.
%% The Appendices part is started with the command \appendix;
%% appendix sections are then done as normal sections
\appendix

%% If you have bib database file and want bibtex to generate the
%% bibitems, please use
%%
\bibliographystyle{elsarticle-num} 
\bibliography{arxiv}

%% else use the following coding to input the bibitems directly in the
%% TeX file.

%% Refer following link for more details about bibliography and citations.
%% https://en.wikibooks.org/wiki/LaTeX/Bibliography_Management
\end{document}